\documentclass[letterpaper, journal, 10 pt, twoside]{IEEEtran}  % Journal final version

\IEEEoverridecommandlockouts                              % This command is only needed if 
\usepackage[english]{babel}
\usepackage[T1]{fontenc}
\usepackage{microtype}
\usepackage{csquotes}

\usepackage{wrapfig}

\usepackage{graphics} % for pdf, bitmapped graphics files
\usepackage{graphicx}
\usepackage{amsmath} % assumes amsmath package installed
\usepackage{bm} % bold for greek letter lowercases, it assumes bm package installed
\usepackage{amssymb}  % assumes amsmath package installed
\usepackage{float}
\usepackage{subfig}
\usepackage{makecell}
\usepackage[normalem]{ulem}
\usepackage{hyperref}
\usepackage{color}
\usepackage{balance}

\usepackage{amsfonts}
\usepackage{algorithm,algpseudocode}
\usepackage[tbtags]{mathtools}

\usepackage[pdftex,dvipsnames]{xcolor}
\usepackage[bottom]{footmisc}		% Force footnotes to be at bottom of the page

\usepackage{booktabs,dcolumn}  %for tables
\newcolumntype{d}[1]{D{.}{.}{#1}} % "decimal" column type

\usepackage{todonotes}
\usepackage{units}

\usepackage{textcomp}
\usepackage{gensymb}

\usepackage[firstpage=true]{background}

\usepackage[comma,numbers,compress]{natbib}

\newcommand{\norm}[1]{\left\lVert#1\right\rVert}

\newcommand\copyrighttext{%
	\parbox{\textwidth}{
		\footnotesize
		\textbf{Accepted final version.} IEEE Robotics and Automation Letters (RA-L), Special Issue on Learning and Control for Autonomous Manipulation Systems: the Role of Dimensionality Reduction. 
		DOI: \href{https://ieeexplore.ieee.org/document/8336974/}{No. 10.1109/LRA.2018.2826057}
	}
}

\SetBgContents{\copyrighttext}
\SetBgScale{1}
\SetBgColor{black}
\SetBgAngle{0}
\SetBgOpacity{1}
\SetBgPosition{current page.south}
\SetBgVshift{0.8cm}
\author{Diego Rodriguez and Sven Behnke% <-this % stops a space
	
\thanks{Manuscript received: November, 21, 2018; Revised January, 26, 2018; Accepted March, 19, 2018.}%Use only for final RAL version
\thanks{This paper was recommended for publication by Editor Han Ding upon evaluation of the Associate Editor and Reviewers' comments.
	This work was supported by the European Union's Horizon 2020 Programme under Grant Agreement 644839 (CENTAURO)
	and the German Research Foundation (DFG) under the grant BE 2556/12 ALROMA in priority programme SPP 1527 Autonomous Learning.} %Use only for final RAL version
\thanks{All authors are with the Autonomous Intelligent Systems (AIS) Group, Computer Science Institute VI, University of Bonn, Germany
	{\tt\small \{rodriguez, behnke\}@ais.uni-bonn.de}}%
\thanks{Digital Object Identifier (DOI): 10.1109/LRA.2018.2826057.}
}

\title{Transferring Category-based Functional Grasping Skills by Latent Space Non-Rigid Registration}

\begin{document}

\maketitle
%\thispagestyle{empty}  
%\pagestyle{empty}

%%%%%%%%%%%%%%%%%%%%%%%%%%%%%%%%%%%%%%%%%%%%%%%%%%%%%%%%%%%%%%%%%%%%%%%%%%%%%%%%
\begin{abstract}
Objects within a category are often similar in their shape and usage.
When we---as humans---want to grasp something, we transfer our knowledge from past experiences and adapt it to novel objects.
In this paper, we propose a new approach for transferring grasping skills that accumulates grasping knowledge into a category-level canonical model.
Grasping motions for novel instances of the category are inferred from geometric deformations between the observed instance and the canonical shape.
Correspondences between the shapes are established by means of a non-rigid registration method that combines the Coherent Point Drift approach with subspace methods.
By incorporating category-level information into the registration, we avoid unlikely shapes and focus on deformations actually observed within the category.
Control poses for generating grasping motions are accumulated in the canonical model from grasping definitions of known objects. 
According to the estimated shape parameters of a novel instance, the control poses are transformed towards it.
The category-level model makes our method particularly relevant for on-line grasping, where fully-observed objects are not easily available.
This is demonstrated through experiments in which objects with occluded handles are successfully grasped.
\end{abstract}

%%%%%%%%%%%%%%%%%%%%%%%%%%%%%%%%%%%%%%%%%%%%%%%%%%%%%%%%%%%%%%%%%%%%%%%%%%%%%%%%
\section{Introduction}
\label{sec:introduction}
\begin{figure}
	\centering
	\includegraphics[width=1\linewidth]{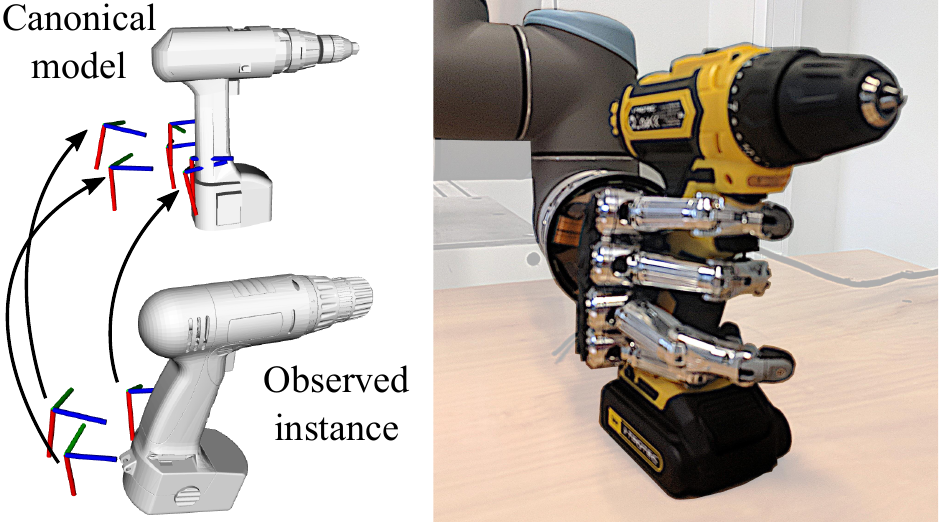}
	\caption{Shape information and grasping knowledge for known object instances are aggregated in a category-level canonical model. Grasping control poses are transferred to novel instances of the category for generating the grasping motion.}
	\label{fig:transfer}
\end{figure}

\IEEEPARstart{W}{hile} transferring grasping skills within a category happens frequently and effortless in humans, obtaining that generalization in robots is still an open problem.  
People can be shown objects that they never saw before, and they often will immediately know how to grasp and operate them.
This happens by transferring knowledge from their learned model of the object category, e.g., screw drivers, to novel instances. 
Although the manipulation of known objects can be planned offline, many open-world applications require the manipulation of unknown instances. 
Our approach accumulates manipulation knowledge of known instances in category-level models and transfers manipulations skills to novel instances (Fig. \ref{fig:transfer}).

The method presented in this paper focuses on functional grasping, i.e., on motions that allow not only to grasp the object but also to use it.
We use the term \emph{grasping} to refer to the process of bringing the object into the hand, and not only to the final configuration of hand and object. 
We propose a method for generating grasping motions for novel instances by making use of category-level  shape information represented by a learned latent shape space.
Our method aggregates object shape and grasping knowledge from multiple known instances of a category in a canonical model. 
The learned latent space of shape variations enables a category-specific shape-aware non-rigid registration procedure that establishes correspondences between a view of a novel object instance and the canonical model. 
Our method finds a transformation from the canonical model to the view in the latent shape space---linearly interpolating and extrapolating from other transformations found within the category---which best matches the observed 3D points.
This estimates the shape parameters of the novel instance and allows for inference of its occluded parts.
By the non-rigid transformation and the aggregated manipulation knowledge, control poses for the novel instance are inferred. 
The grasping motion is finally generated by using those control poses.

In this paper, we extend our previous work \cite{Rodriguez2017} by accumulating grasping knowledge in the canonical model in addition to the shape information, which enriches our transferring skill model.

\section{Related Work}
\label{sec:related_work}
\subsection{Non-Rigid Registration and Shape Spaces}
\label{sec:Registration}
Most of the non-rigid registration methods proposed so far differ mostly by the prior restrictions or regularization on the deformation that the points can undergo. 
Several restrictions such as conformal maps \citep{levy2002least, zeng2010dense, kim2011blended}, isometry \citep{bronstein2006efficient, tevs2009isometric, ovsjanikov2010one}, thin-plate splines \citep{allen2003space, brown2007global}, elasticity \citep{haehnel2003extension} and Motion Coherence Theory \citep{myronenko2010point} have been used to encourage or constrain different types of transformations. 

For surface reconstruction, many methods use non-rigid registration \citep{li2009robust, sussmuth2008reconstructing, wand2009efficient, newcombe2015dynamicfusion}. 
Approaches such as presented by \citet{li2009robust, zollhofer2014real} sequentially add higher frequency details coming from new depth camera frames to a low-resolution 3D capture through non-rigid registration.

For category-based shape spaces, several methods have been proposed.
\citet{hasler2009statistical} generate a shape space of human bodies with poses using 3D markers and human scans.
\citet{burghard2013compact} developed a shape space of varying geometry based on dense correspondences.
\citet{engelmann2016joint} define a shape manifold which models intra-class shape variance; this method is robust with noisy or occluded regions. 

\subsection{Transferring Grasping Skills}
Based on segmented objects according to their RGB-D appearance, \citet{vahrenkamp} transfer grasp poses from a set of template grasps.
\citet{ficuciello} developed an approach to confer grasping capabilities based on a reinforcement learning technique and postural synergies.
In \citep{stouraitis} and \citep{hillenbrand}, functional grasp poses are warped such that distance between correspondences is minimized, then the warped poses are replanned in order to increase the functionality of the grasp.
In \citep{amor} a similar contact warping is combined with motor synergies to generalize human grasping.
\citet{stuckler2011real} transfer manipulation skills using a non-rigid registration method based on multi-resolution surfel maps.
The non-rigid registration serves as the mechanism to warp available grasping poses.

\subsection{Discussion}
\label{sec:Discussion}
Although current state-of-the-art methods for non-rigid registration yield good results, they have some limitations.
\citet{newcombe2015dynamicfusion} use optical flow constraints and thus this approach does not perform well with large deformations or changes in color and illumination.
Moreover, several captures of the object are required.
The method by \citet{burghard2013compact} accurately estimates dense correspondences, but does not perform well with incomplete scans or noisy data.
To solve these problems, we incorporate category-level information in our approach, such that we are able to register partially-occluded novel instances using a single capture of the object.
Methods such as \citet{engelmann2016joint} deal with minor misalignments and occlusions, but do not offer correspondences between points and do not give any kind of transformation.
Our method, on the other hand, offers a transformation for each point of the novel instance and even points that do not belong to the object which allows us to transform grasp poses.

Regarding transferring grasping skills, we tackle the problem of requiring a fully observed \citep{stouraitis} or a non-occluded \citep{vahrenkamp} object by exploiting the geometrical information residing in our learned categorical model.
Unlike \citep{stuckler2011real} we model shape and grasping not for single known instances, but for object categories, which gives us the possibility to learn typical shape variations and to infer grasping information even when parts of the object are not observed.
More importantly, none of previous approaches is able to accumulate and to use knowledge from \emph{several} previous successfully experiences, which is the main focus of this paper.
\section{Method}
\label{sec:method}

\begin{figure*}
	\centering
	\includegraphics[width=1.0\linewidth]{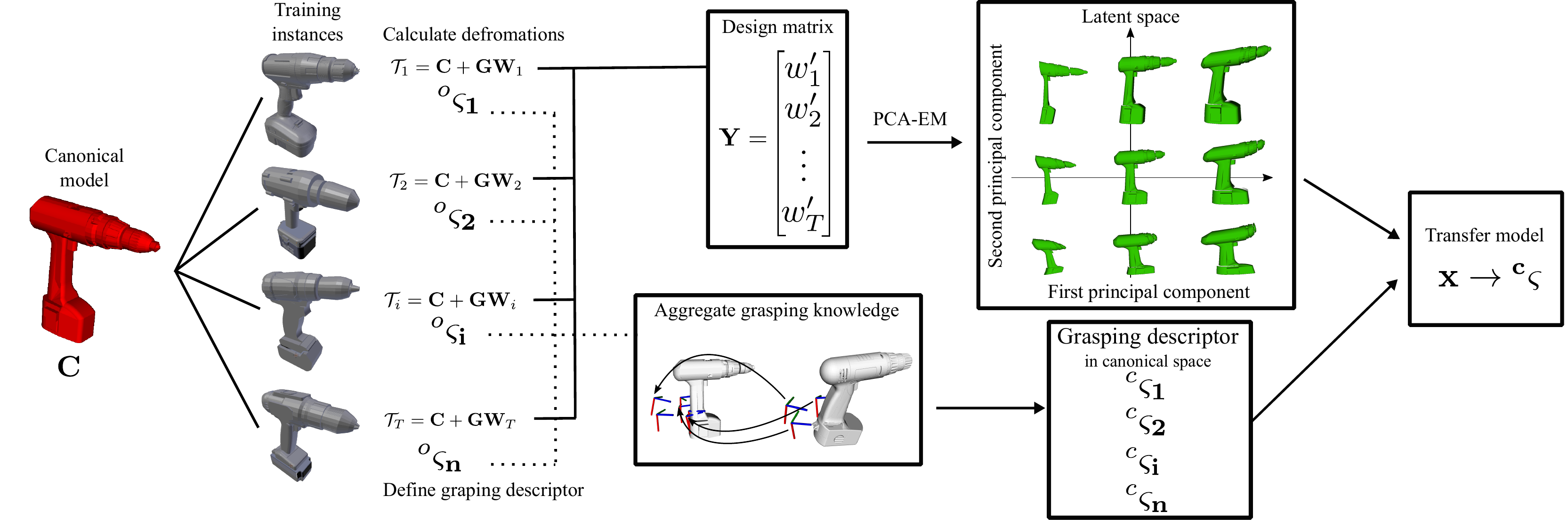} 
	\caption{Training phase. The deformations between each instance and the canonical model are calculated using CPD. These deformations are assembled into the design matrix $\mathbf{Y}$. Using PCA-EM, the principal components which constitute the latent space are extracted. The grasping descriptor for each training sample is aggregated in the canonical model. The latent variables serve as feature vector while the grasping descriptor is the desired output for the grasping transfer model.}
	\label{fig:overview}
\end{figure*}

\begin{figure*}
	\centering
	\includegraphics[width=1.0\linewidth]{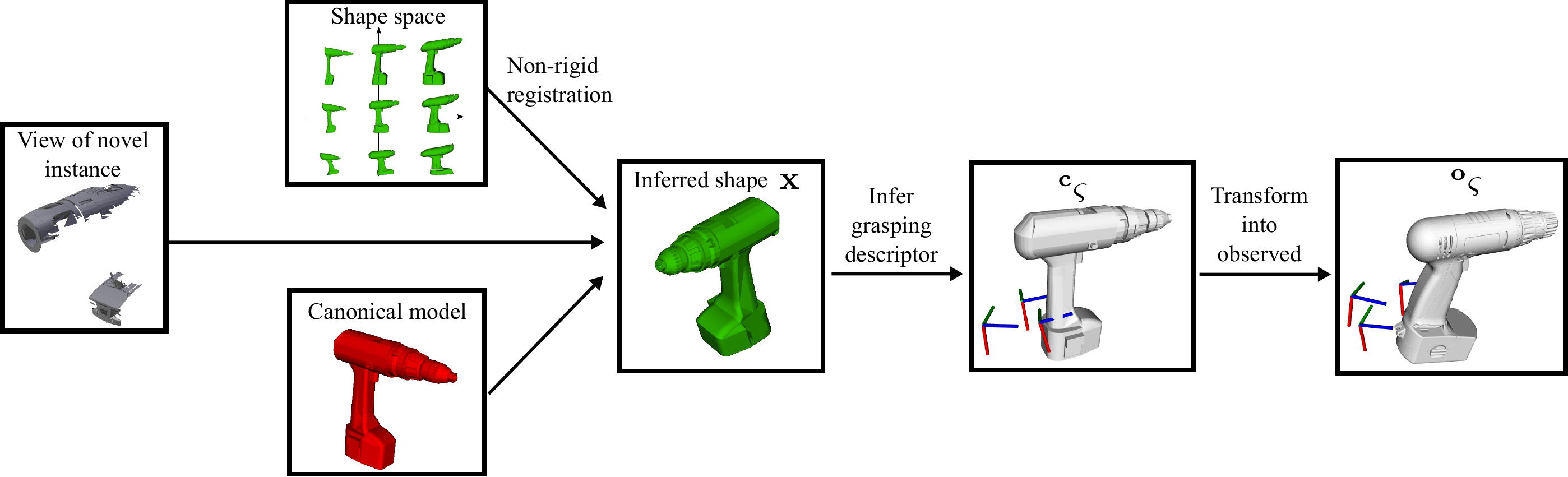}
	\caption{The canonical shape (red) is matched against a partially-occluded target shape (leftmost) by finding its latent shape parameters. The grasping descriptor $^{c}\bm{\varsigma}$ is inferred from $\mathbf{x}$. Finally, the descriptor is transformed to the observed space.}
	\label{fig:inference}
\end{figure*}

Our approach is composed of a learning phase and an inference phase (Figs.~\ref{fig:overview} and \ref{fig:inference}). 
In the learning phase, a category-specific linear model of the transformations that a category of objects can undergo is built. 
In this manner, poses in the space of the canonical shape can be transformed into the space of an observed instance. 
These poses can be added even after the learning phase.
The category-specific linear model is learned as follows: First, we select a single instance from the training dataset to be the canonical model of the category. Then, we find the transformations relating this instance to all other instances of the category using Coherent Point Drift (CPD) \citep{myronenko2010point}. Finally, we find a linear latent subspace of these transformations, which becomes our transformation model for the category. 
For each instance in the training set, an associated grasping descriptor $\bm{\varsigma}$ (vector representation of the grasping motion) is also transformed into the canonical space.
In this manner, multiple experiences can be aggregated in the canonical model.

In the inference phase, given a novel observed instance, our method searches in the subspace of transformations to find the transformation which best relates the canonical shape to the observed instance. 
Depending on the resulting latent shape variables and the aggregated grasping knowledge accumulated in the canonical model, a grasping descriptor for the novel instance is inferred.

\subsection{Categories and Shape Representation}
\label{sec:Classes_and_Shape_Representation}
A category is composed by a set of objects which share the same topology and have a similar shape. 
Each category has a canonical shape $\mathbf{C}$ that will be deformed to fit the shape of the training and testing sample shapes.
To represent a shape, we use point clouds, which can be generated from meshes by ray-casting from several viewpoints on a tessellated sphere and then down-sampling with a voxel grid filter. 
Each category specifies a canonical pose and reference frame, used for initial alignments. 

\subsection{Coherent Point Drift}
\label{sec:cpd}
Here, we shortly describe the Coherent Point Drift (CPD) \citep{myronenko2010point} and how we use it for our non-rigid registration.

CPD estimates a deformation field mapping between a template point set $\mathbf{S}^{[t]}=(\mathbf{s}^{[t]}_1, ..., \mathbf{s}^{[t]}_M)^T$ and a reference point set $\mathbf{S}^{[r]}=(\mathbf{s}^{[r]}_1, ..., \mathbf{s}^{[r]}_N)^T$.
The points in $\mathbf{S}^{[t]}$ are modeled as centroids of a Gaussian Mixture Model (GMM) from which the points in $\mathbf{S}^{[r]}$ are drawn. 
CPD maximizes the likelihood of the GMM while imposing constraints on the motion of the centroids such that points near each other should move coherently and have a similar motion to their neighbors \citep{yuille1988motion}.
The likelihood of the GMM is not directly maximized, but instead its equivalent negative log-likelihood function is minimized:
\begin{equation}
    \label{eq:Deformation_Energy}
    E(\bm{\psi},\sigma^2) = -\sum^N_{n=1}{\log{\sum^M_{m=1}{\exp^{-\frac{1}{2\sigma^2} \norm{\mathbf{s}^{[r]}_n-\mathcal{T}(\mathbf{s}^{[t]}_m, \bm{\psi})}^2}}}},
\end{equation}
where $\mathcal{T}(\mathbf{s}^{[t]}_m, \bm{\psi})$ is a parametrized transformation from the template point set to the reference set, and $\sigma^2$ is the covariance of the Gaussian density.
The transformation $\mathcal{T}$, for the non-rigid registration, is defined as the initial position plus a displacement function $v$: 
\begin{equation}
    \label{eq:Deformation_Vector}
    \mathcal{T}(\mathbf{S}^{[t]},v) = \mathbf{S}^{[t]} + v(\mathbf{S}^{[t]}).
\end{equation}
The constraints on the motion of the centroids are realized by regularizing the displacement function $v$. Adding this regularization $\phi(v)$ to the negative log-likelihood Eq. (\ref{eq:Deformation_Energy}), we obtain
\begin{equation}
\label{eq:e+regularization}
f(v, \sigma^2) = E(\sigma^2, v)+\frac{\lambda}{2}\phi(v),
\end{equation}
where $\lambda$ is a trade-off parameter between the goodness of maximum likelihood fit and regularization. A particular choice of $\phi(v)$ leads to the following displacement function $v(\mathbf{Z})$~\citep{myronenko2010point}:
\begin{equation}
    \label{eq:Deformation_Field}
    v(\mathbf{Z}) = G(\mathbf{S^{[t]}}, \mathbf{Z})\mathbf{W},
\end{equation}
for any set of $D$-dimensional points $\mathbf{Z}_{N\times D}$.
$G(\mathbf{S^{[t]}}, \mathbf{Z})$ is defined as a Gaussian kernel matrix composed element-wise by:
\begin{equation}
    \label{eq:Gaussian_Kernel}
	g_{ij} = G(\mathbf{s}^{[t]}_i, \mathbf{z}_j) = \exp^{ -\frac{1}{2\beta^2} \norm{ \mathbf{s}^{[t]}_i-\mathbf{z}_j} ^2},
\end{equation}
$\mathbf{W}_{M\times D}$ is a matrix of kernel weights, 
and $\beta$ is a scalar that controls the strength of interaction between points. 
An additional interpretation of $\mathbf{W}$ is as a set of $D$-dimensional deformation vectors, each associated with one of the $M$ points of $\mathbf{S^{[t]}}$.
For convenience in the notation, $\mathbf{G}_{M\times M}$ will denote $G(\mathbf{S}^{[t]},\mathbf{S}^{[t]})$.
Note that $G(\cdot,\cdot)$ can simply be computed by Eq. (\ref{eq:Gaussian_Kernel}), but the matrix $\mathbf{W}$ needs to be estimated. 

To minimize Eq. (\ref{eq:e+regularization}), CPD uses an Expectation Maximization (EM) algorithm.
In the E-step, the posterior probabilities matrix $\mathbf{P}$ is estimated using past parameter values. 
%To add robustness to outliers, an additional uniform probability distribution is added to the mixture model. 
This matrix $\mathbf{P}$ is composed element-wise by:
\begin{equation}
    \label{eq:E-Step}
    p_{mn} = \frac{e^{-\frac{1}{2\sigma^2}\norm{\mathbf{s}^{[r]}_n-(\mathbf{s}^{[t]}_m+G(m,\cdot)\mathbf{W})}^2}} {\sum^M_{m=1}{e^{-\frac{1}{2\sigma^2}\norm{\mathbf{s}^{[r]}_n-(\mathbf{s}^{[t]}_m+G(k,\cdot)\mathbf{W})}^2}}+\frac{\omega}{1-\omega}\frac{(2\pi\sigma^2)^{\frac{D}{2}}}{N}}
\end{equation}
where $\omega$ reflects the assumption on the amount of noise.

In the M-step, the matrix $\mathbf{W}$ is estimated by:
\begin{equation}
    \label{eq:M-Step}
    (\mathbf{G}+\lambda\sigma^2d(\mathbf{P1})^{-1})\mathbf{W} = d(\mathbf{P1})^{-1}\mathbf{P}\mathbf{S}^{[r]} - \mathbf{S}^{[t]}
\end{equation}
where $\mathbf{1}$ represents a column vector of ones and $d(\cdot)^{-1}$ is the inverse diagonal matrix. For a more detailed description of the CPD algorithm, please refer to~\citep{myronenko2010point}.

In our method, we use the canonical shape $\mathbf{C}$ for the deforming template shape $\mathbf{S^{[t]}}$ and each training example $\mathbf{T}_i$ as the reference point set $\mathbf{S^{[r]}}$. Therefore, the transformations $\mathcal{T}_i$ are defined as 
\begin{equation}
\label{eq:defomation_field}
\mathcal{T}_i(\mathbf{C},\mathbf{W}_i) = \mathbf{C} + \mathbf{G}\mathbf{W}_i
\end{equation}

where $\mathbf{W}_i$ is the $\mathbf{W}$ matrix computed by taking training example $\mathbf{T}_i$ as the reference point set $\mathbf{S^{[r]}}$.

\subsection{Latent Space}
CPD allows us to define a feature vector representing the deformation field. 
This vector has the same length for all training examples; additionally, elements in this vector correspond with the same elements in another. 
This allows us to learn a latent lower-dimensional space.

We observe from Eq. (\ref{eq:defomation_field}) that the deformation field between the canonical and an observed instance is fully determined by $\mathbf{G}$ and $\mathbf{W}$.
Moreover, we see that $\mathbf{G}$ only requires the points of the canonical shape and it remains constant for all training examples. 
Therefore, the entire uniqueness of the deformation field for each training example is captured by its matrix $\mathbf{W}$.

We construct a row vector $\mathbf{y}_i\in \mathbb{R}^{p=M\cdot D}$ from each matrix $\mathbf{W}_i$ of each training example $\mathbf{T}_i$, that characterizes the corresponding deformation field. 
The vectors are normalized to have zero-mean and unit-variance and are then assembled into a design matrix $\mathbf{Y}$. 
Finally, we find a lower-dimensional manifold of deformation fields for the category by applying the Principle Component Analysis Expectation Maximization (PCA-EM) algorithm on the matrix $\mathbf{Y}$. 

Much like with CPD, we alternate between an E- and M-step. The E-step is given by:
\begin{equation}
\label{PCA_E-Step}
\mathbf{X} = \mathbf{Y}\mathbf{L}^T(\mathbf{L}\mathbf{L}^T)^{-1}
\end{equation}
whereas the M-step is defined by:
\begin{equation}
\label{PCA_M-Step}
\mathbf{L} = (\mathbf{X}^T\mathbf{X})^{-1}\mathbf{X}^T\mathbf{Y}.
\end{equation}
$\mathbf{L}_{p \times q}$ is the resulting matrix of principle components.
So, for a new normalized set of observations $\mathbf{Y_o}$, the latent variables can be found by postmultiplying $\mathbf{Y_o}$ by $\mathbf{L}$.
In this manner, a deformation field is now described by only $q$ latent parameters.
Similarly, any point $\mathbf{x}$ in the latent space can be converted into a deformation field transformation by first postmultiplying $\mathbf{x}$ by $\mathbf{L}^T$ and by converting the result into a $\mathbf{W}_{M \times D}$ matrix after the respective denormalization.
Thus, moving through the $q$-dimensional space linearly interpolates between the deformation fields.

\subsection{Grasping Knowledge Aggregation}
\label{sec:grasp_aggregation}

We aggregate grasping knowledge from different instances into the canonical model in two steps: first, by generating the grasping motion in the observed space and, second, by transforming its grasping descriptor into the canonical space.

A grasping motion is represented as a sequence of parametrized primitives each of them defined by a control pose expressed in the same coordinate system of the shape of the object.
The generation of grasping motions can be performed manually for each instance in the training set, which favors accuracy over time and wear off of the system (on real robotic platforms). 
This imposes however a limit on the number of samples of the training dataset mostly because of time constraints. 
In order to overcome this limit, we adopt a constrained sample-based motion generation approach.

\begin{figure}[b]
	\centering
	\includegraphics[width=0.9\linewidth]{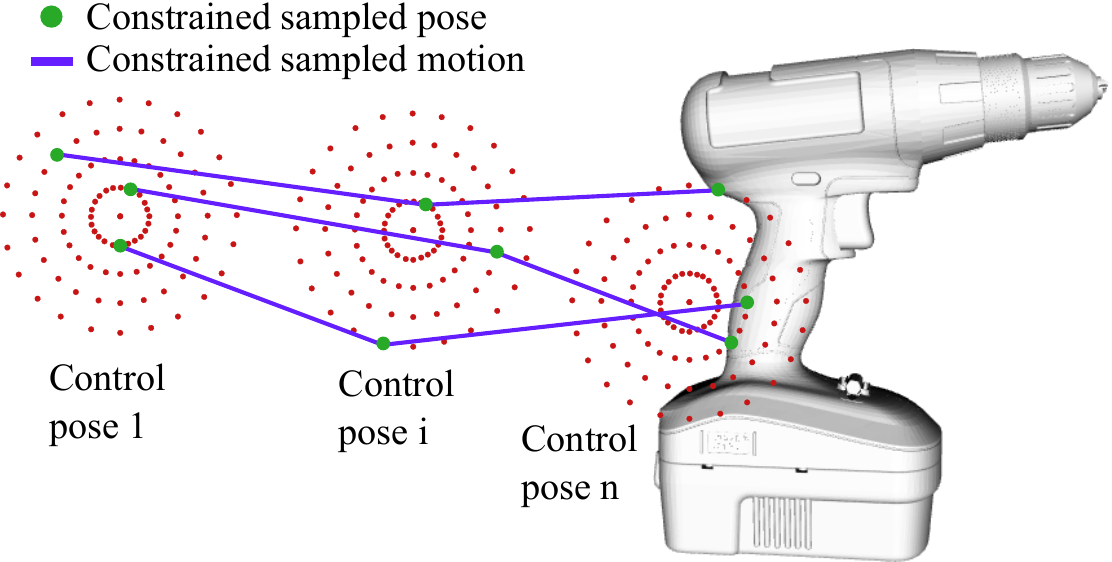} 
	\caption{Sampled-based grasping motion generation. 6D constrained random poses are sampled around control poses of the canonical grasping motion.}
	\label{fig:motion_sampling}
\end{figure}

A sampled motion is created by generating constrained random 6D poses around the control poses of the canonical grasping motion as depicted in Figure \ref{fig:motion_sampling}.
Each component of the translation is sampled from a normal distribution. 
For the rotation, a quaternion is build out of three uniformed points following the approach described in \cite{Shoemake}. 
These orientations are filtered by specific functional constraints of each category, in the case of drills, for example, rotations that occlude or impede the use of the trigger are discarded.
If the sampled grasping motion leads to collisions with other objects in the environment including the robotic arm, the motion is discarded as well.
Finally, the sampled motion is executed and evaluated.
If the object is functionally grasped successfully, the grasping control poses are transformed into the canonical space.

Finding the transformation from the observed space into the canonical space is equal to finding the inverse transformation of Eq. (\ref{eq:Deformation_Vector}) or equivalently to finding the inverse transformation of Eq. (\ref{eq:Deformation_Field}). 
However, the inverse function $v^{-1}$ is not directly available.
It can nonetheless be estimated for a point $\mathbf{o}$ in the space of the observed shape using a set of points $\mathbf{Z} = (\mathbf{z}_1, ..., \mathbf{z}_M)^T$ in the canonical space which deform close to $\mathbf{o}$ by the equation:
\begin{equation}
\label{eq:Inverse_Deformation}
v^{-1}(\mathbf{o}) = -\frac{\sum_{i=1}^M{G(\mathbf{o}, \mathbf{z}_i + v(\mathbf{z}_i))v(\mathbf{z}_i)}}{\sum_{i=1}^M{G(\mathbf{o}, \mathbf{z}_i + v(\mathbf{z}_i))}}.
\end{equation}

For transforming the orientation, we apply Eq. (\ref{eq:Inverse_Deformation}) to the rotational vector base of each pose and orthonormalize it.

\begin{figure}[b]
	\centering
	\includegraphics[width=0.8\linewidth]{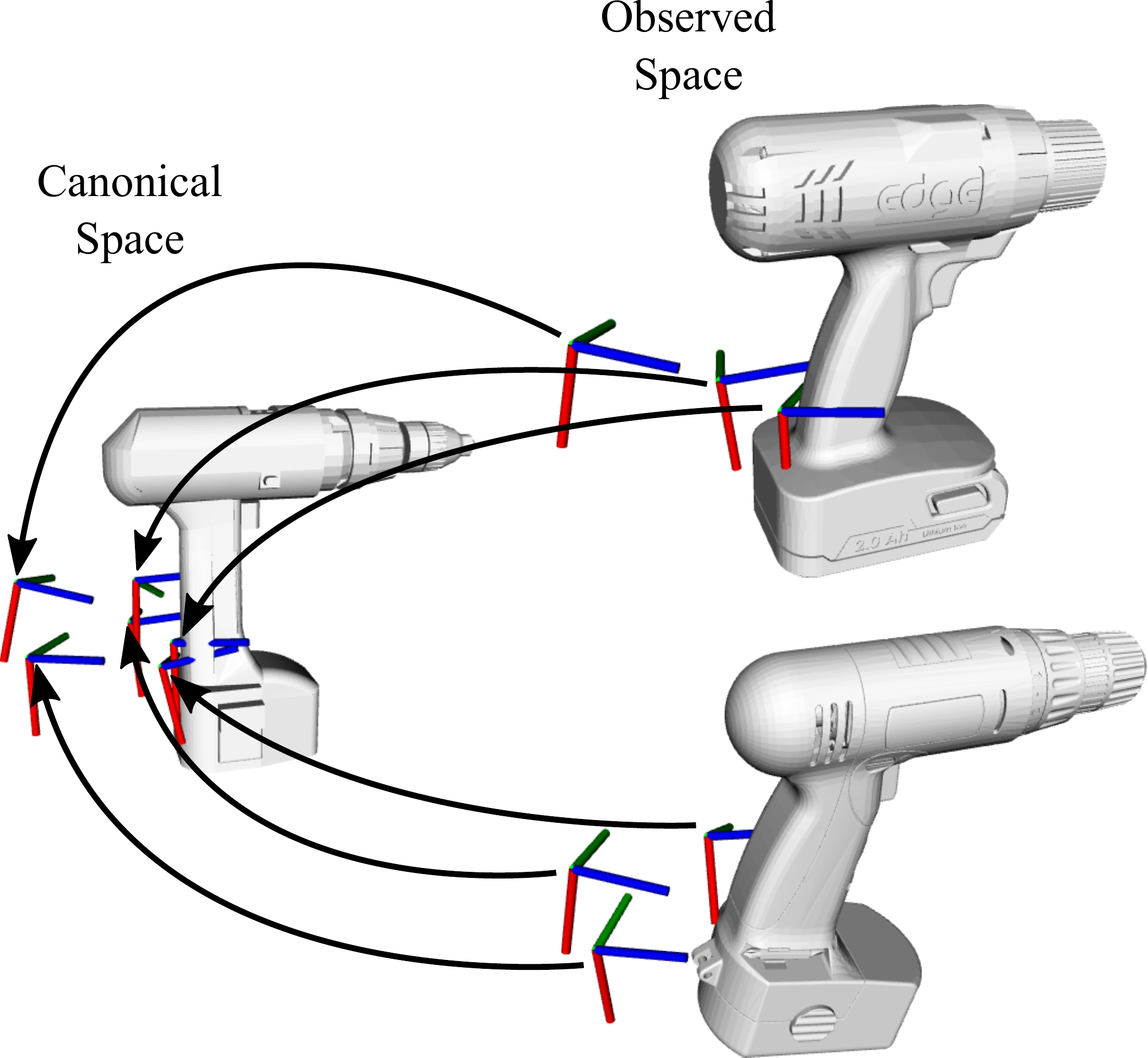} 
	\caption{Grasping knowledge aggregation. Grasping descriptors of observed instances are transformed and aggregated in the canonical model by Eq. (\ref{eq:Inverse_Deformation})}
	\label{fig:aggregation}
\end{figure}

For each instance in our training dataset, we have so far a latent vector $\mathbf{x}_i$ that represents the shape deformations from the canonical instance to the observed instance and a grasping descriptor $\bm{\varsigma}_i$ transformed into the canonical space. 
We set the latent vector $\mathbf{x}_i$ as a feature vector and the grasping descriptor $\bm{\varsigma}_i$ as the corresponding target output and train a linear regression model. 
In other words, grasping knowledge is aggregated in the canonical model by serving as a training label of a regression model (Fig. \ref{fig:aggregation}).
Algorithm \ref{alg:Training} summarizes the training phase (Figure \ref{fig:overview}).

\subsection{Shape Inference}
\label{sec:Inference}
A shape transformation is specified by the $q$ parameters of the latent vector $\mathbf{x}_i$ plus additional seven parameters of a rigid transformation $\bm{\theta}_i$. 
The rigid transformation is meant to account for minor misalignments between the observed shape and the canonical shape at the global level. 
\begin{algorithm}[b]
	\caption{Training phase}
	\label{alg:Training}
	\textbf{Input:} A set of training shapes in their canonical pose with corresponding grasping descriptors $^{o}\bm{\varsigma}$.
	\begin{algorithmic}[1]
		\State Select a canonical shape $\mathbf{C}$ via heuristic or \-pick \-the one with the lower reconstruction energy.
		\State Estimate the deformation fields between the ca\-no\-ni\-cal shape and the other training examples using CPD.
		\State Concatenate the resulting set of $\mathbf{W}$ matrices from the deformation fields into a design matrix $\mathbf{Y}$.
		\State Perform PCA-EM on the design matrix $\mathbf{Y}$ to compute the latent space of deformation fields $\mathbf{x}$.
		\State Transform the grasping descriptors $^{o}\bm{\varsigma}$ into the canonical space $^{c}\bm{\varsigma}$.
		\item Train the Linear Regressor $\mathcal{R}:\mathbf{x}\rightarrow^{c}\bm{\varsigma}$.
	\end{algorithmic}
	\textbf{Output:} A canonical shape $\mathbf{C}$, a latent space of deformation fields $\mathbf{L}$ and a trained model for inferring grasping descriptors $\mathcal{R}$.
\end{algorithm}

We concurrently optimize for the latent parameters and the rigid transformation using gradient descent. 
As CPD and ICP, our method requires an initial coarse alignment of the observed shape because of the expected local minima.  
We want to find an aligned dense deformation field which when applied to the canonical shape $\mathbf{C}$ minimizes the distance to corresponding points in the observed shape $\mathbf{O}$. 
Specifically, we want to minimize the energy function:
\begin{equation}
\label{eq:Energy}
E(\mathbf{x},\bm{\theta}) = -\sum^{M}_{m=1}{\log{\sum^{N}_{n=1}{\exp^{\frac{1}{2\sigma^2}\norm{\mathbf{O}_n-\Theta(\mathcal{T}_m(\mathbf{C}_m,\mathbf{W}_m(\mathbf{x})),\bm{\theta})}^2}}}}
\end{equation}
where the function $\Theta$ applies the rigid transformation given parameters $\bm{\theta}$.

When a minimum is found, we can transform any point or set of points into the observed space by applying the deformation field using Eq. (\ref{eq:Deformation_Field}) and Eq. (\ref{eq:Deformation_Vector}) and then applying the rigid transformation $\Theta$.
Moreover, CPD provides a dense deformation field, allowing us to find deformation vectors for novel points, even those added after the field is created. 

\subsection{Transferring Grasping Skills}
The transfer of grasping skills for novel instances is performed as follows. 
A latent vector $\mathbf{x}$ describing the shape deformation of the object from the canonical instance is calculated as explained in Section \ref{sec:Inference}.
This vector constitutes a test sample of the linear regression, whose inference is a grasping descriptor $^{c}\bm{\varsigma}$.
Then, $^{c}\bm{\varsigma}$ is transformed into the observed space.
This transformation is performed in two steps. 
First, the control poses of the grasping motion are warped using Eq. (\ref{eq:Deformation_Vector}) replacing $\mathbf{S}^{[t]}$ by the translational part and the rotational vector base of the control poses. 
Because the warping process can violate the orthogonality of the orientation, we orthonormalize the warped orientation.
Second, we apply the rigid transformation $\Theta$ defined by the parameters $\bm{\theta}$.

The resulting transformed control poses $^{o}\bm{\varsigma}$ are expressed in the frame of the object. 
Thus, for executing the motion each of the poses has to be adapted relative to the pose of the observed object by premultiplying the control poses by the pose of the object w.r.t. the base of the manipulator. Algorithm \ref{alg:Inference} summarizes the inference of grasping skills. 

\begin{algorithm}
	\caption{Grasping Skills Inference}
	\label{alg:Inference}
	\textbf{Input:} Transformation model ($\mathbf{C}$, $\mathbf{L}$), trained regressor $\mathcal{R}$ and observed shape $\mathbf{O}$
	\begin{algorithmic}[1]
		\State Use gradient descent to estimate the parameters of the underlying transformation ($\mathbf{x}$ and $\bm{\theta}$) until the termination criteria is met. To calculate the value of the energy function, in each iteration: 
		
		-  Using the current values of $\mathbf{x}$ and $\bm{\theta}$:
		\begin{enumerate}
			\item Create vector $\mathbf{\hat{Y}}$ and convert it into matrix $\mathbf{W}$.
			\item Use Eq. (\ref{eq:Deformation_Field}) and Eq. (\ref{eq:Deformation_Vector}) to deform $\mathbf{C}$.
			\item Apply the rigid transformation $\Theta$ to the deformed $\mathbf{C}$.
		\end{enumerate}
		\State Use the resulting $\mathbf{x}$ to infer a grasping descriptor $^{c}\bm{\varsigma}$ inferred by $\mathcal{R}$.
		\State Transform the grasping descriptor into the observed space. 
	\end{algorithmic}
	\textbf{Output:} Grasping descriptor in observed space $^{c}\bm{\varsigma}$.
\end{algorithm}

\section{Setup and Evaluation}
\label{sec:experiments}
In this section, we evaluate only the grasping skill transfer because the latent space non-rigid registration method was already evaluated in \cite{Rodriguez2017}.
We tested our method on two categories: \textit{Drill} a \textit{Spray Bottle}, containing $13$ and $17$ instances respectively. 
We obtained the object models from two online CAD databases: GrabCad \footnote{\url{https://grabcad.com/library}} and 3DWarehouse\footnote{\url{https://3dwarehouse.sketchup.com/}}. 
The CAD models were converted into meshes in order to generate the input point clouds for our method. 
They were obtained by ray-casting from several viewpoints on a tessellated sphere and down-sampling with a voxel grid filter.

\begin{figure}[b]
	\centering
	\includegraphics[width=1.0\linewidth]{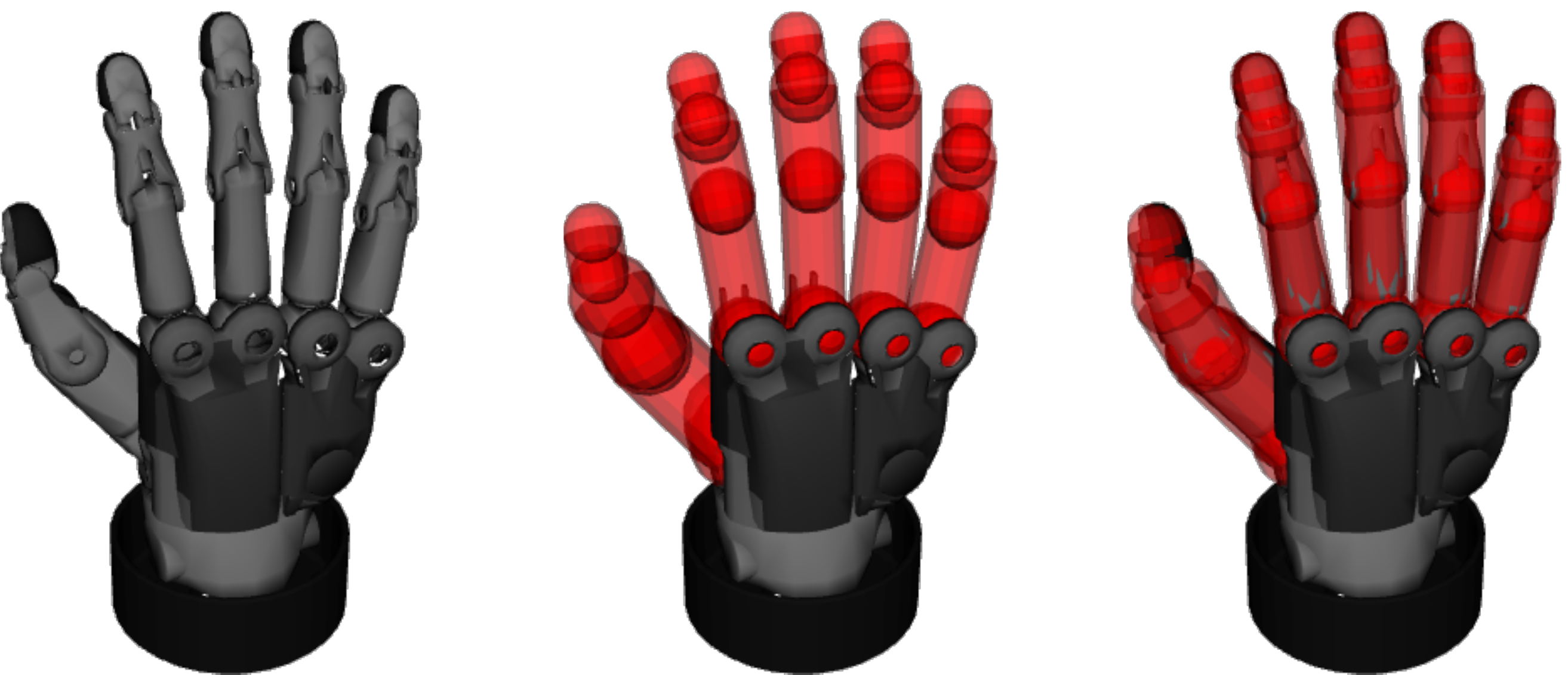}
	\caption{Visual and collision model of the robotic hand. At rightmost both models are displayed simultaneously to show the goodness of the capsule approximation.}
	\label{fig:collisions}
\end{figure}

We use the five-fingered Schunk hand with a total of $9$ fully actuated Degrees of Freedom (DoF) and $11$ mimic joints. 
The experiments were carried out in the Gazebo simulation environment.
The collision model of the finger links were modeled by capsules using an automatic ROS optimal capsule generator based on the Roboptim library \cite{Khoury} as shown in Fig. \ref{fig:collisions}.
The inertia tensors of the graspable objects were approximated using Meshlab.
For building the shape latent spaces, we parametrized CPD with $\beta\!=\!1$, $\lambda\!=\!3$ and $\sigma^2=0.01$. 
The number of latent variables was set to capture at least $95\%$ of the variance of each class.
The grasping motions for each object in the training set were sampled as described in Section~\ref{sec:grasp_aggregation} with a maximum distance of \unit[0.04]{m} and a maximum angular deviation of $0.2$.

For each category, we select the canonical model manually. We use cross validation leaving two samples out.
We trained six \textit{drill} and seven \textit{spray bottle} grasping transfer models.
Because our method is able to infer category-alike geometries, we also evaluated our method with partially-observed point clouds.
For this, we generate a single view of the test objects of each cross validation model.
In total, we evaluated the method on $12$ fully observed and $12$ partially observed \textit{drills} and $14$ fully observed and $14$ partially observed spray bottles.
For each instance, one simulation trial was performed because the execution of the generated motion is fully deterministic in simulation.

From the $52$ instances to be grasped $30$ were successfully grasped; that yields a success rate of $57.7\%$. 
Note, however, that a successfully grasped instance in our approach considers the entire motion, not only the last grasp configuration.

\begin{table}[t]
	\captionsetup{justification=centering, font={sc, small}, labelsep=newline,singlelinecheck=false}
	\caption{Ratio of successfully transfered grasps.}
	\begin{tabular*}{\linewidth}{@{\extracolsep{\fill}} l *{4}{c} }
		\toprule
		& \multicolumn{2}{c}{Drill} & \multicolumn{2}{c}{Spray Bottle}\\
		& \multicolumn{1}{c}{Grasp} & \multicolumn{1}{c}{Func. Grasp} & \multicolumn{1}{c}{Grasp} & \multicolumn{1}{c}{Func. Grasp}\\
		\midrule
		Fully observed		& 7/12 & 4/12 & 8/14 & 3/14 \\
		Partially observed 	& 6/12 & 3/12 & 9/14 & 6/14 \\
		\bottomrule
	\end{tabular*}
	\label{tab:results_experiments}
\end{table}

\begin{figure*}
	\includegraphics[width=1.0\linewidth]{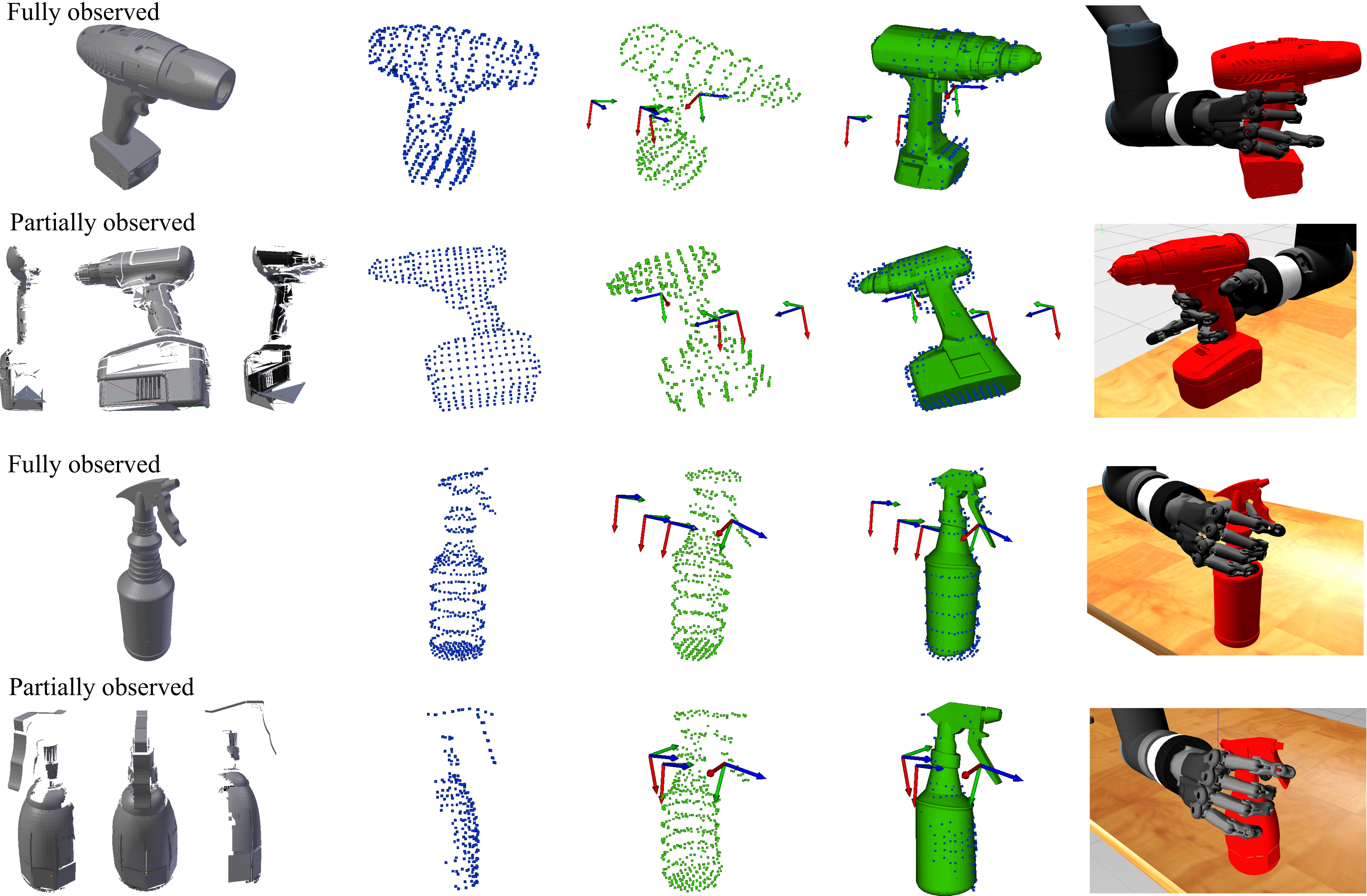}
	\caption{At the leftmost the meshes are shown.
		For illustration purposes we show additional perspectives of the same single view of the partially observed objects.
		The respective point clouds are shown in blue.
		The inferred instances (green point clouds) together with the transformed control points that define the motion are also displayed.
		In order to observe how good the inference matches the observed points, the mesh of the canonical models is transformed and displayed (green meshes) together with the observed data (blue points).
		Finally, the resulting grasped object in Gazebo is also depicted at the rightmost.}
	\label{fig:experiments} 
\end{figure*}

Regarding functional grasps, i.e., the index finger is able to trigger the tools, $16$ instances were successfully grasped which results in a $32\%$ success rate.
The results are presented in Table \ref{tab:results_experiments}.
Compared to the results presented in \cite{stouraitis}, although the success rate of our method is lower, our method is able to handle partially-occluded objects and an inference takes in average \unit[7]{s} compared to the \unit[12.6]{min} which is only suitable for offline applications.
Figure \ref{fig:experiments} shows for each category two different---a fully observed and a partially-observed---samples that were successfully grasped.

\begin{figure*}
	\centering
	\includegraphics[width=0.3\linewidth]{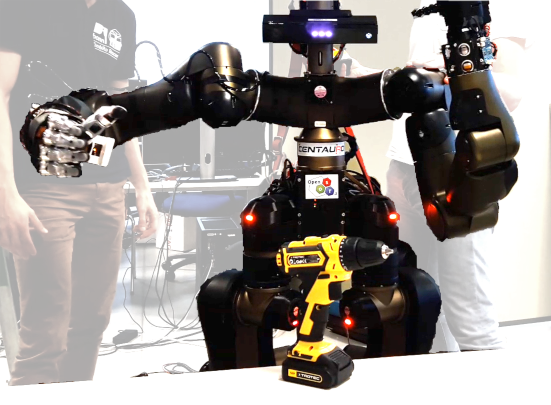}
	\includegraphics[width=0.28\linewidth]{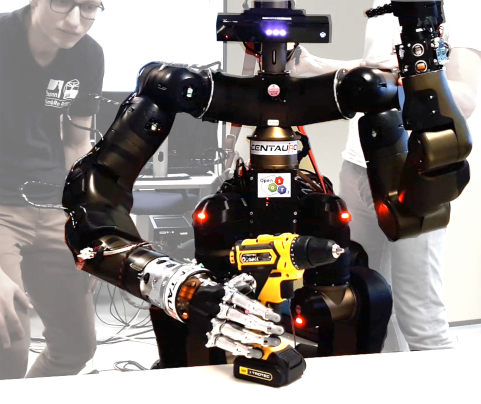} 
	\includegraphics[width=0.3\linewidth]{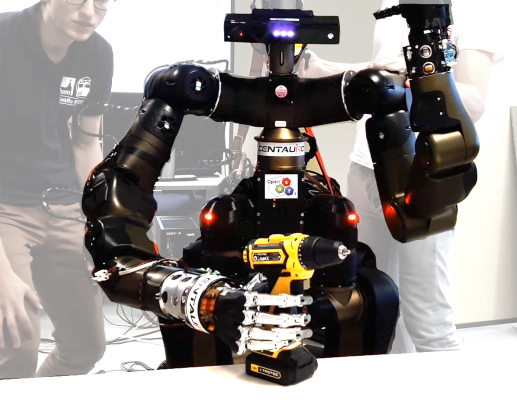} 
	\includegraphics[width=0.3\linewidth]{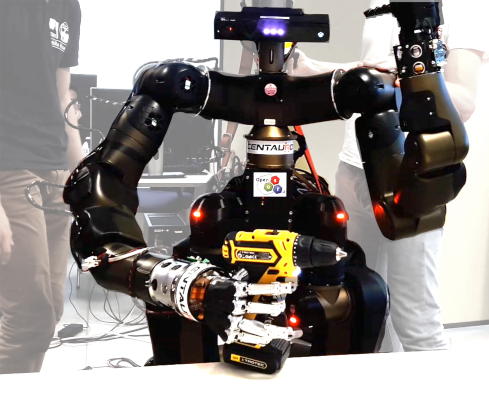} 
	\includegraphics[width=0.3\linewidth]{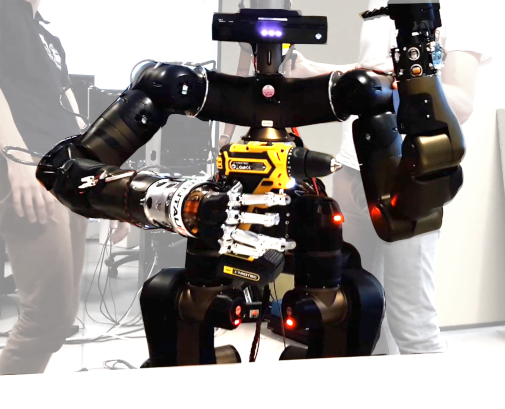} 
	\caption{Experiments performed with the Centauro robot grasping autonomously a novel instance of a drill.}
	\label{fig:centauro_experiments}
\end{figure*}

Our method was also tested in real-robot experiments. 
We created only one latent transformation model for the \textit{drill} category using all the $12$ available meshes plus the canonical model. 
The observed object was inferred from one single view captured by the Kinect v2 sensor~\cite{fankhauser2015}.
The tests were carried out on two different platforms: a UR10 arm and the CENTAURO robot.
The hand was controlled by a PID position-current cascade controller, such that the joint position controller defines the desired joint currents.
The saturation values of the current controller together with the PID values of the position controller were set to provide a certain level of compliance which contributed mainly at the last stage of the grasping motion. 
Using the UR10 robotic arm, our method was able to grasp two different drills twice without any failure. 
Similarly, with the CENTAURO robot, our approach grasped one instance of a drill twice without any failure (Fig. \ref{fig:centauro_experiments}).

A video illustrating our approach is available
online\footnote{\url{http://www.ais.uni-bonn.de/videos/RA-L_2018_Rodriguez}}.

\subsection{Discussion}
\label{sec:exp:discussion}
Real experiments with two different robotic arms demonstrate that our method does not depend on the kinematics of the arm holding the hand.
We assume however that the kinematics of the arm is able to reach 6D poses in its workspace.
Our method is also agnostic to the robotic hand; a canonical grasping motion that is suitable to the hand is the only requirement for applicability.

Most of the grasping motions that failed exhibited a high deviation with respect to the canonical control poses which indicates a large variance in the learned transfer model.
This suggests a need for more sample-efficient inference methods and the need for more training data.

\section{Conclusion}
\label{sec:conclusion}
In this paper, we proposed a new approach of transferring grasping skills between objects within a category that is based on the knowledge aggregation of different training samples into a canonical model.
Thanks to the learned latent shape space, our method is capable of completing missing or occluded object surfaces from partial views. 
Our method was able to transfer grasping skills with real robotic platforms from experiences collected only in simulation.
This demonstrates the feasibility regarding the available sensory data (single-view point clouds) and runtime of our approach.

For future work, we want to consider more complex categories that impose higher variations in the joint configuration of the hand.
So, more dimensionality reduction will be expected.
As we realized the reduced number of training samples limits the presented approach, we start looking into automatic generation of plausible meshes from the canonical model.
We also want to explore variants of the CPD algorithm in order to speed our current implementation.
Finally, we would like also to exploit additional sensory modalities such as joint currents and force-torque sensors.

\balance

%\printbibliography

\bibliographystyle{IEEEtranN}
\bibliography{RA-L_grasp2018}

% Generated by IEEEtranN.bst, version: 1.14 (2015/08/26)
\begin{thebibliography}{29}
\providecommand{\natexlab}[1]{#1}
\providecommand{\url}[1]{#1}
\csname url@samestyle\endcsname
\providecommand{\newblock}{\relax}
\providecommand{\bibinfo}[2]{#2}
\providecommand{\BIBentrySTDinterwordspacing}{\spaceskip=0pt\relax}
\providecommand{\BIBentryALTinterwordstretchfactor}{4}
\providecommand{\BIBentryALTinterwordspacing}{\spaceskip=\fontdimen2\font plus
\BIBentryALTinterwordstretchfactor\fontdimen3\font minus
  \fontdimen4\font\relax}
\providecommand{\BIBforeignlanguage}[2]{{%
\expandafter\ifx\csname l@#1\endcsname\relax
\typeout{** WARNING: IEEEtranN.bst: No hyphenation pattern has been}%
\typeout{** loaded for the language `#1'. Using the pattern for}%
\typeout{** the default language instead.}%
\else
\language=\csname l@#1\endcsname
\fi
#2}}
\providecommand{\BIBdecl}{\relax}
\BIBdecl

\bibitem[Rodriguez et~al.(2018)Rodriguez, Cogswell, Koo, and
  Behnke]{Rodriguez2017}
D.~Rodriguez, C.~Cogswell, S.~Koo, and S.~Behnke, ``Transferring grasping
  skills to novel instances by latent space non-rigid registration,'' in
  \emph{IEEE International Conference on Robotics and Automation (ICRA)}, 2018.

\bibitem[L{\'e}vy et~al.(2002)L{\'e}vy, Petitjean, Ray, and
  Maillot]{levy2002least}
B.~L{\'e}vy, S.~Petitjean, N.~Ray, and J.~Maillot, ``Least squares conformal
  maps for automatic texture atlas generation,'' in \emph{{ACM} Transactions on
  Graphics (TOG)}, vol.~21, no.~3, 2002, pp. 362--371.

\bibitem[Zeng et~al.()Zeng, Wang, Wang, Gu, Samaras, and
  Paragios]{zeng2010dense}
Y.~Zeng, C.~Wang, Y.~Wang, X.~Gu, D.~Samaras, and N.~Paragios, ``Dense
  non-rigid surface registration using high-order graph matching,'' in
  \emph{Computer Vision and Pattern Recognition (CVPR), 2010 IEEE Conference
  on}, pp. 382--389.

\bibitem[Kim et~al.(2011)Kim, Lipman, and Funkhouser]{kim2011blended}
V.~G. Kim, Y.~Lipman, and T.~Funkhouser, ``Blended intrinsic maps,'' in
  \emph{{ACM} Transactions on Graphics (TOG)}, vol.~30, no.~4, 2011, p.~79.

\bibitem[Bronstein et~al.(2006)Bronstein, Bronstein, and
  Kimmel]{bronstein2006efficient}
A.~M. Bronstein, M.~M. Bronstein, and R.~Kimmel, ``Efficient computation of
  isometry-invariant distances between surfaces,'' \emph{SIAM Journal on
  Scientific Computing}, vol.~28, no.~5, pp. 1812--1836, 2006.

\bibitem[Tevs et~al.()Tevs, Bokeloh, Wand, Schilling, and
  Seidel]{tevs2009isometric}
A.~Tevs, M.~Bokeloh, M.~Wand, A.~Schilling, and H.-P. Seidel, ``Isometric
  registration of ambiguous and partial data,'' in \emph{Computer Vision and
  Pattern Recognition (CVPR), 2009 IEEE Conference on}, pp. 1185--1192.

\bibitem[Ovsjanikov et~al.(2010)Ovsjanikov, M{\'e}rigot, M{\'e}moli, and
  Guibas]{ovsjanikov2010one}
M.~Ovsjanikov, Q.~M{\'e}rigot, F.~M{\'e}moli, and L.~Guibas, ``One point
  isometric matching with the heat kernel,'' in \emph{Computer Graphics Forum},
  vol.~29, no.~5.\hskip 1em plus 0.5em minus 0.4em\relax Wiley Online Library,
  2010, pp. 1555--1564.

\bibitem[Allen et~al.(2003)Allen, Curless, and Popovi{\'c}]{allen2003space}
B.~Allen, B.~Curless, and Z.~Popovi{\'c}, ``{The space of human body shapes:
  reconstruction and parameterization from range scans},'' in \emph{{ACM}
  Transactions on Graphics ({TOG})}, vol.~22, no.~3, 2003, pp. 587--594.

\bibitem[Brown and Rusinkiewicz(2007)]{brown2007global}
B.~J. Brown and S.~Rusinkiewicz, ``Global non-rigid alignment of 3-d scans,''
  in \emph{{ACM} Transactions on Graphics (TOG)}, vol.~26, no.~3, 2007, p.~21.

\bibitem[Hahnel et~al.(2003)Hahnel, Thrun, and Burgard]{haehnel2003extension}
D.~Hahnel, S.~Thrun, and W.~Burgard, ``An extension of the {ICP} algorithm for
  modeling nonrigid objects with mobile robots,'' in \emph{18th International
  Joint Conference on Artificial Intelligence (IJCAI)}, 2003, pp. 915--920.

\bibitem[Myronenko and Song(2010)]{myronenko2010point}
A.~Myronenko and X.~Song, ``Point set registration: {C}oherent point drift,''
  \emph{IEEE Transactions on Pattern Analysis and Machine Intelligence (PAMI)},
  vol.~32, no.~12, pp. 2262--2275, 2010.

\bibitem[Li et~al.(2009)Li, Adams, Guibas, and Pauly]{li2009robust}
H.~Li, B.~Adams, L.~J. Guibas, and M.~Pauly, ``Robust single-view geometry and
  motion reconstruction,'' in \emph{{ACM} Transactions on Graphics (TOG)},
  vol.~28, no.~5, 2009, p. 175.

\bibitem[S{\"u}{\ss}muth et~al.(2008)S{\"u}{\ss}muth, Winter, and
  Greiner]{sussmuth2008reconstructing}
J.~S{\"u}{\ss}muth, M.~Winter, and G.~Greiner, ``Reconstructing animated meshes
  from time-varying point clouds,'' in \emph{Computer Graphics Forum}, vol.~27,
  no.~5.\hskip 1em plus 0.5em minus 0.4em\relax Wiley Online Library, 2008, pp.
  1469--1476.

\bibitem[Wand et~al.(2009)Wand, Adams, Ovsjanikov, Berner, Bokeloh, Jenke,
  Guibas, Seidel, and Schilling]{wand2009efficient}
M.~Wand, B.~Adams, M.~Ovsjanikov, A.~Berner, M.~Bokeloh, P.~Jenke, L.~Guibas,
  H.-P. Seidel, and A.~Schilling, ``Efficient reconstruction of nonrigid shape
  and motion from real-time {3D} scanner data,'' \emph{{ACM} Transactions on
  Graphics (TOG)}, vol.~28, no.~2, p.~15, 2009.

\bibitem[Newcombe et~al.(2015)Newcombe, Fox, and
  Seitz]{newcombe2015dynamicfusion}
R.~A. Newcombe, D.~Fox, and S.~M. Seitz, ``Dynamicfusion: {R}econstruction and
  tracking of non-rigid scenes in real-time,'' in \emph{IEEE Conference on
  Computer Vision and Pattern Recognition (CVPR)}, 2015, pp. 343--352.

\bibitem[Zollh{\"o}fer et~al.(2014)Zollh{\"o}fer, Nie{\ss}ner, Izadi, Rehmann,
  Zach, Fisher, Wu, Fitzgibbon, Loop, Theobalt, et~al.]{zollhofer2014real}
M.~Zollh{\"o}fer, M.~Nie{\ss}ner, S.~Izadi, C.~Rehmann, C.~Zach, M.~Fisher,
  C.~Wu, A.~Fitzgibbon, C.~Loop, C.~Theobalt \emph{et~al.}, ``Real-time
  non-rigid reconstruction using an {RGB-D} camera,'' \emph{{ACM} Transactions
  on Graphics (TOG)}, vol.~33, no.~4, p. 156, 2014.

\bibitem[Hasler et~al.(2009)Hasler, Stoll, Sunkel, Rosenhahn, and
  Seidel]{hasler2009statistical}
N.~Hasler, C.~Stoll, M.~Sunkel, B.~Rosenhahn, and H.-P. Seidel, ``A statistical
  model of human pose and body shape,'' in \emph{Computer Graphics Forum},
  vol.~28, no.~2.\hskip 1em plus 0.5em minus 0.4em\relax Wiley Online Library,
  2009, pp. 337--346.

\bibitem[Burghard et~al.(2013)Burghard, Berner, Wand, Mitra, Seidel, and
  Klein]{burghard2013compact}
O.~Burghard, A.~Berner, M.~Wand, N.~Mitra, H.-P. Seidel, and R.~Klein,
  ``Compact part-based shape spaces for dense correspondences,'' \emph{CoRR},
  vol. abs/1311.7535, 2013.

\bibitem[Engelmann et~al.(2016)Engelmann, St{\"u}ckler, and
  Leibe]{engelmann2016joint}
F.~Engelmann, J.~St{\"u}ckler, and B.~Leibe, ``Joint object pose estimation and
  shape reconstruction in urban street scenes using {3D} shape priors,'' in
  \emph{German Conference on Pattern Recognition (GCPR)}, 2016, pp. 219--230.

\bibitem[Vahrenkamp et~al.(2016)Vahrenkamp, Westkamp, Yamanobe, Aksoy, and
  Asfour]{vahrenkamp}
N.~Vahrenkamp, L.~Westkamp, N.~Yamanobe, E.~E. Aksoy, and T.~Asfour,
  ``Part-based grasp planning for familiar objects,'' in \emph{16th IEEE-RAS
  International Conference on Humanoid Robots (Humanoids)}, 2016, pp. 919--925.

\bibitem[Ficuciello et~al.()Ficuciello, Zaccara, and Siciliano]{ficuciello}
F.~Ficuciello, D.~Zaccara, and B.~Siciliano, ``Synergy-based policy improvement
  with path integrals for anthropomorphic hands,'' in \emph{2016 IEEE/RSJ
  International Conference on Intelligent Robots and Systems (IROS)}, pp.
  1940--1945.

\bibitem[Stouraitis et~al.()Stouraitis, Hillenbrand, and Roa]{stouraitis}
T.~Stouraitis, U.~Hillenbrand, and M.~A. Roa, ``Functional power grasps
  transferred through warping and replanning,'' in \emph{2015 IEEE
  International Conference on Robotics and Automation (ICRA)}, pp. 4933--4940.

\bibitem[Hillenbrand and Roa(2012)]{hillenbrand}
U.~Hillenbrand and M.~A. Roa, ``Transferring functional grasps through contact
  warping and local replanning,'' in \emph{IEEE/RSJ International Conference on
  Intelligent Robots and Systems (IROS)}, 2012, pp. 2963--2970.

\bibitem[Amor et~al.(2012)Amor, Kroemer, Hillenbrand, Neumann, and
  Peters]{amor}
H.~B. Amor, O.~Kroemer, U.~Hillenbrand, G.~Neumann, and J.~Peters,
  ``Generalization of human grasping for multi-fingered robot hands,'' in
  \emph{IEEE/RSJ International Conference on Intelligent Robots and Systems
  (IROS)}, 2012, pp. 2043--2050.

\bibitem[Stueckler et~al.(2011)Stueckler, Steffens, Holz, and
  Behnke]{stuckler2011real}
J.~Stueckler, R.~Steffens, D.~Holz, and S.~Behnke, ``Real-time {3D} perception
  and efficient grasp planning for everyday manipulation tasks.'' in
  \emph{European Conference on Mobile Robots (ECMR)}, 2011, pp. 177--182.

\bibitem[Yuille and Grzywacz(1988)]{yuille1988motion}
A.~L. Yuille and N.~M. Grzywacz, ``The motion coherence theory,'' in
  \emph{Computer Vision, 2nd International Conference on (ICCV)}.\hskip 1em
  plus 0.5em minus 0.4em\relax IEEE, 1988, pp. 344--353.

\bibitem[Shoemake(1992)]{Shoemake}
K.~Shoemake, ``Uniform random rotations,'' in \emph{Graphics Gems}.\hskip 1em
  plus 0.5em minus 0.4em\relax Morgan Kaufmann, 1992, pp. 124--132.

\bibitem[Khoury et~al.()Khoury, Lamiraux, and Taix]{Khoury}
A.~E. Khoury, F.~Lamiraux, and M.~Taix, ``Optimal motion planning for humanoid
  robots,'' in \emph{2013 IEEE International Conference on Robotics and
  Automation (ICRA)}, pp. 3136--3141.

\bibitem[Fankhauser et~al.(2015)Fankhauser, Bloesch, Rodriguez, Kaestner,
  Hutter, and Siegwart]{fankhauser2015}
P.~Fankhauser, M.~Bloesch, D.~Rodriguez, R.~Kaestner, M.~Hutter, and
  R.~Siegwart, ``Kinect v2 for mobile robot navigation: Evaluation and
  modeling,'' in \emph{International Conference on Advanced Robotics (ICAR)},
  2015, pp. 388--394.

\end{thebibliography}

\end{document}